\documentclass[iicol,sn-aps]{sn-jnl}

\usepackage{graphicx}%
\usepackage{multirow}%
\usepackage{amsmath,amssymb,amsfonts}%
\usepackage{amsthm}%
\usepackage{mathrsfs}%
\usepackage[title]{appendix}%
\usepackage{xcolor}%
\usepackage{textcomp}%
\usepackage{manyfoot}%
\usepackage{booktabs}%
\usepackage{algorithm}%
\usepackage{algorithmicx}%
\usepackage{algpseudocode}%
\usepackage{listings}%
\usepackage{bm}

\usepackage{booktabs, tabularx}
\usepackage{makecell}

\pagecolor{white}

\def\R{\mathbb{R}}

\DeclareMathOperator{\softmax}{softmax}

\newcommand{\frm}[1]{\langle #1\rangle}

\newcommand{\revv}[1]{#1}

\begin{document}

\title[Humanising robot-assisted navigation]{Humanising robot-assisted navigation}

\author*[1]{\fnm{Placido} \sur{Falqueto}} \email{placido.falqueto@unitn.it}
\author[1]{\fnm{Alessandro} \sur{Antonucci}} \email{alessandro.antonucci@unitn.it}
\author[1]{\fnm{Luigi} \sur{Palopoli}} \email{luigi.palopoli@unitn.it}
\author[2]{\fnm{Daniele} \sur{Fontanelli}} \email{daniele.fontanelli@unitn.it}
\affil*[1]{\orgdiv{Department of Information Engineering and Computer Science (DISI)}, \orgname{University of Trento}, \orgaddress{\street{Via Sommarive 9}, \city{Trento}, \postcode{38123}, \country{Italy}}}
\affil[2]{\orgdiv{Department of Industrial Engineering (DII)}, \orgname{University of Trento}, \orgaddress{\street{Via Sommarive 9}, \city{Trento}, \postcode{38123}, \country{Italy}}}

\keywords{Shared Control, Human-Centered Robotics, Motion and Path Planning, Physically Assistive Devices}

\abstract{
    Robot-assisted navigation is a perfect example of a class of
    applications requiring flexible control approaches. When the human
    is reliable, the robot should concede space to their initiative.
    When the human makes inappropriate choices the robot controller
    should kick-in guiding them towards safer paths. Shared authority
    control is a way to achieve this behaviour by deciding online how
    much of the authority should be given to the human and how much
    should be retained by the robot. An open problem is how to
    evaluate the appropriateness of the human’s choices. One possible
    way is to consider the deviation from an ideal path computed by
    the robot. This choice is certainly safe and efficient, but it
    emphasises the importance of the robot’s decision and relegates
    the human to a secondary role. In this paper, we propose a
    different paradigm: a human’s behaviour is correct if, at every
    time, it bears a close resemblance to what other humans do in
    similar situations. This idea is implemented through the
    combination of machine learning and adaptive control. \revv{The
    map of the environment is decomposed into a grid. In each cell, we
    classify the possible motions that the human executes. We use a
    neural network classifier to classify the current motion, and the
    probability score is used as a hyperparameter in the control to
    vary the amount of intervention}. The experiments collected for
    the paper show the feasibility of the idea. \revv{A qualitative
    evaluation, done by surveying the users after they have tested the
    robot, shows that the participants preferred our control method
    over a state-of-the-art visco-elastic control}.
}
  
\maketitle



\section{Introduction}
\label{sec:introduction}

We are living a time when robots are no longer confined to industrial
environments but are used in numerous applications that require an
unprecedented degree of autonomy. Modern robots have to adapt to
humans, to understand their needs and help them carry out their
activities.  
It is easy to predict a future in which the interaction between robots
and humans will draw a direct inspiration from the rider-horse
metaphor~\cite{flemisch2003h}, with the human having developed an
``innate'' ability to use the robot's services, and the robot being
able to grasp the human's intention without any explicit request.

\begin{figure}[t!]
	\centering
	\begin{tabularx}{0.5\textwidth} { >{\centering\arraybackslash}X
   >{\centering\arraybackslash}X  }
   \includegraphics[width=\linewidth]{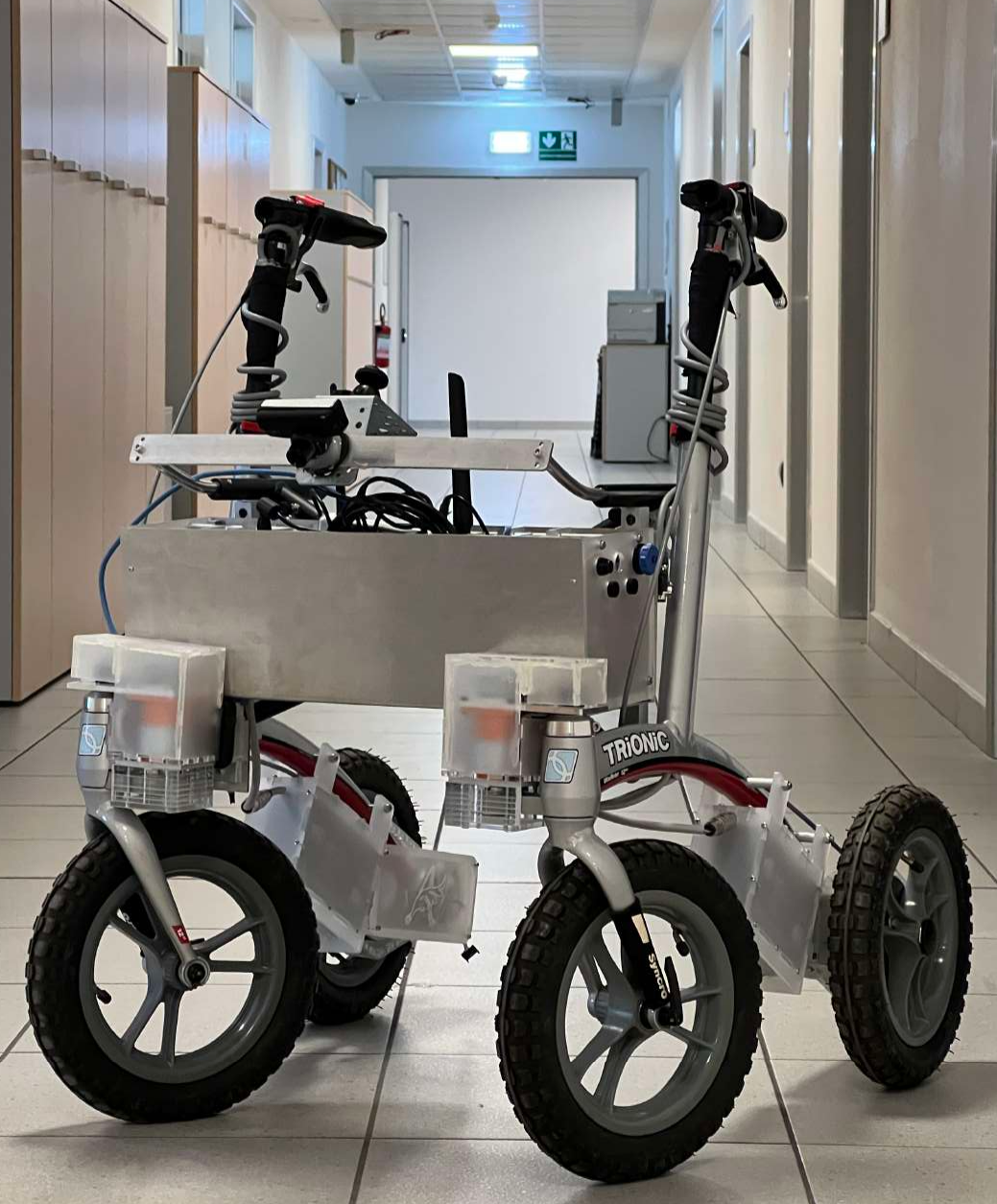}
   & \includegraphics[width=\linewidth]{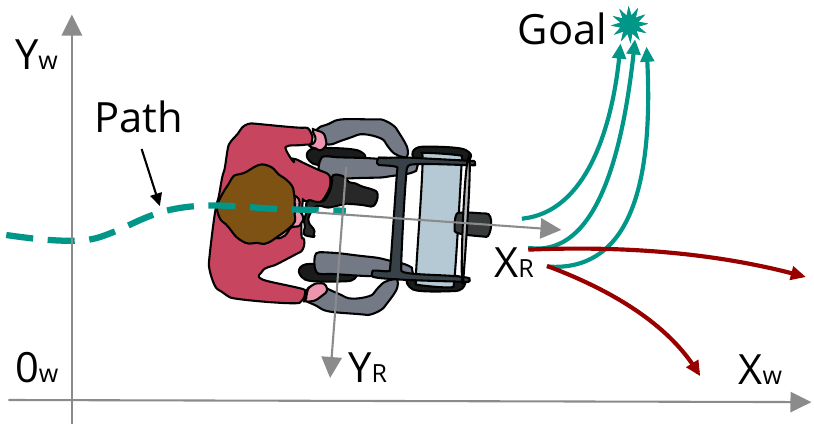}
   \\
		(a) & (b) \\
	\end{tabularx}
	\caption{(a) The {\em FriWalk} used in our experiments. (b) Walker
		representation and adopted reference frames $\frm{W}$ and
		$\frm{R}$. In the example shown, the robot will allow any
		movement to the left (green curves), while it will perform a
		corrective action for all the non compliant ones (red
		curves).}
	\label{fig:representation-walker}
\end{figure}

In this paper we consider a use case in which the {\em FriWalk}
robotic rollator (see Fig.~\ref{fig:representation-walker}-a), a
robotic assistant able to localise itself in complex environments and
generate safe routes, is used as navigation and walking support for a
person with mild cognitive deficits.  \revv{We consider the case of
users with mild cognitive deficits, who could find it difficult to
plan and follow long paths across complex and populated environments.
These conditions could generate stress and fatigue and determine a
gradual withdrawal of the user from the public space. A moderate
cognitive support implemented by a system that gently guides the user
can play a useful role in recovering or preserving the user's sense of
direction and her/his cognitive abilities.} Since our ultimate goal is
to prolong and promote the human's autonomy, the system seeks to leave
the human in control of the guidance as much as possible taking over
the trajectory control only when the human's actions are evidently
flawed. This is a specific paradigm of a more general idea dubbed
\textit{shared authority control}, in which the robot is moved partly
by the human and partly by the autonomous system, based on the
contingent situation. In our previous work~\cite{andreetto2017path},
we developed a guidance system of this kind based on a hybrid control
scheme. In that study, the system leaves the user in control or kicks
in the automatic guidance based on safety considerations. This is a
remarkable simplification of the design space: the amount of authority
reserved to the guidance system is increased when the distance from
the border of a safe virtual corridor becomes too narrow. Similarly
in~\cite{MagnagoADFP18icra} the shift in authority is based on the
localisation accuracy. The reference trajectory can be found by the
use of optimal path planning algorithms, even for dynamic
scenarios~\cite{bevilacqua2016path}.

The previous paradigms used a robot-centric and, to some extent,
patronising point-of-view: the ideal behaviour is one where the human
would do exactly the same things that the robot has planned through
optimal path planning.
In this paper we increase the level of our ambition: we aim to take
away control from the human only in presence of erratic behaviours
that deviate from what any other \emph{human} would do in the same
situation. \revv{At the same time, we aim to give the highest possible
level of freedom to the user, without forcing her/him to follow a
particular trajectory.} This leads us to a central question: how do
people move when they travel between two destinations in a known
environment? The human body is extremely versatile and can generate
innumerable motion patterns. However, in the past few years we learned
that when humans walk, they tend to minimise the derivative of the
curvature, which is a quantity related with the
jerk~\cite{arechavaleta2008optimality}; this implies that they follow
regular motion patterns similar to those generated by a nonholonomic
vehicle.  These trajectories are well approximated by clothoids
\revv{(formal definition of a clothoid curve in
Section~\ref{subsec:map-generation})}.  Still, different individuals
generate different classes of trajectories. In terms of navigation, we
can intuitively acknowledge that there are different ways to ``turn
right'', which are equally permissible and that together share the
attribute of being a ``right turn''.  Starting from this observation,
we have developed a learning-based framework to classify the features
of human motion from synthetically generated trajectories, \revv{based
on the geometric properties of the paths,} thus creating a grid of
expected behaviours. \revv{We define the minimalistic set \{Left-turn,
Right-turn, Straight\} to identify the behaviour.} Given a navigation
task, we look for the behaviour that more closely matches the current
motion in each portion of the traversed space. The motion control
calculates the likelihood with which the current motion belongs to the
current region of interest. When the likelihood becomes too low, it
means that the human is behaving unexpectedly and the controller
shifts the shared authority towards the robot. This way, the human is
left in control as long as they perform as other people would do in
the same environment. \revv{If the user is strongly committed to
follow a path different from the one suggested by the system, s/he can
force her way. When the system detects a strong opposition, it is
disengaged for a configurable time unless a strong safety risk is
detected (e.g., the presence of stairways).}

From the implementation point of view, the walker is actuated with a
visco--elastic control that regulates the steering angle. The
classification confidence on the observed features is used as
hyper-parameter to determine the visco--elastic force.  The paradigm
is illustrated in Fig.~\ref{fig:representation-walker}-b.

The paper is organised as follows. In Section~\ref{sec:related}, we
summarise the most important literature contributions used as
reference in this work. In Section~\ref{sec:problem} we formally
describe the state of the problem, while in Section~\ref{sec:model}
our proposed framework is presented in detail. The simulation and
experimental results on the robot are reported in
Section~\ref{sec:experiments}, and finally in
Section~\ref{sec:conclusions} we give our conclusions and announce
future work directions.


\section{Related work}
\label{sec:related}

This paper presents strong elements of novelty, which are supported by
important previous research results in several areas that will be
discussed in this section.

In this work we use a robotic walker which supports a person during
locomotion. As described in~\cite{roro2019} a robotic walker can have
multiple functions to support different problems in the elderly
locomotion. The robotic rollator can support the human's mobility,
increase safety and self-empowerment, compensate unbalanced gait, aid
during the sitting or getting up phases and be used for
rehabilitation~\cite{palopoli2015ISR} \cite{lopes2016}.

When robots need to move or to guide someone mimicking the human
motion, it is essential to have a model of human motion. It is rather
established that humans actually move following smooth
trajectories~\cite{arechavaleta2008nonholonomic} and that their motion
model is well approximated by a unicycle~\cite{farina2017walking},
which naturally generates clothoid curves, also known as Euler
spirals.  As well as being used to express human-like
motion~\cite{antonucci2021humans}, clothoids have important
properties: they are smooth, have a linear curvature and can be
expressed through a simple and analytic form that makes them easy to
manage in real--time implementations~\cite{bertolazzi2018g2}.
Arechavaleta et al.~\cite{arechavaleta2008optimality} use a dynamic
extension of the unicycle model and a numerical optimisation algorithm
to find optimal solutions that well approximate the human locomotor
trajectories. The cost functional minimises the time derivative of the
curvature when the linear velocity is assumed to be constant and
positive. 

In our previous work~\cite{AntonucciPBPF21access} the human motion is
successfully predicted using a neural network to learn the parameters
of the Social Force Model, a physics-based model which describes the
motion of people in social contexts, considering the person as a
particle subject to attractive forces and repulsive forces. In works
such as~\cite{Ziebart}, \cite{Kuderer} and \cite{Kretzschmar} a
combination of Inverse Reinforcement Learning (IRL) and the principle
of maximum entropy was used to learn pedestrian decision making
protocols from large volumes of data.

Different works investigated how to retrieve abstract information
about the behaviour of humans from their trajectories, the majority of
which use machine learning.  For instance, Support Vector Machines
(SVM) have been used to classify different walking styles and
behaviours, including movements such as straight, left-turn,
right-turn, U-turn, and not walking~\cite{kanda2009abstracting}. The
input samples included the normalised coordinates, the orientation,
the velocities, and the bounding boxes of the trajectories over the
observation window. Instead, with Autoencoders (AEs), it is possible
to reconstruct the data elaborated by the system while learning lower
dimensional representations of the data, referred usually as its {\em
  latent space} representation. A combination of clustering and linear
AEs was proposed by~\cite{murray2020dual} to predict the future
trajectory of vessels. The vessel state space is comprised of pose and
linear velocities. In~\cite{rakos2020compression} a convolutional
Variational Autoencoder (VAE) was used to train a latent
representation of real-world vehicle trajectories, represented as a
time series of 2D coordinates; the authors claim a reduction of the
dimension of the latent space from 10 to 2, with evident benefits on
the reconstruction ability without evident reduction of the
classification accuracy. Although it is commonly believed that one of
the goals of modern machine learning is to identify useful
characteristics from simple time series of the coordinates, we argue
that some prior geometrical information can be easily extracted and
used as input for the neural model, in order to simplify its
development and to improve the accuracy of the results. In a similar
way, Lu et al.~\cite{lu2020dual}, augment the input of a Convolutional
Autoencoder (CAE) with spatio-temporal information such as velocity,
acceleration, and the heading change rate.  \revv{ As explained
  in~\cite{bank2023autoencoders}, the encoder part of an autoencoder
  can be used to extract the features of the input and can be connected
  to a classification network to achieve classification
  abilities. This is exactly what we do in this work to classify the
  human trajectories. To the best of our knowledge there is no past
  work that uses the structure {\tt encoder + classification network}
  to determine to which class the analysed trajectory pertains to
  (Left-turn, Right-turn, Straight).}

\revv{ As discussed below, we use an autoencoder network to extract
  the geometric parameters of each trajectory, and a second neural
  network to associate the geometric parameters with a class of
  behaviours.  While the classification problem could be solved by
  other means, the use of learning approaches has two significant
  advantages: first, the use of NNs defines a general framework that
  can be easily generalised to other types of geometric features and
  behaviour classes; second, at runtime the classification produced by
  the NN can be used to produce a score that quantifies the degree of
  agreement between observed and expected behaviour. Specifically,}
during the control phase, we use a Bayesian technique to extract a
measure of confidence on the behaviour currently followed by the
human.

\revv{To the best of our knowledge, this paper is the first to use the
  neural network's confidence score to control the amount of
  intervention in a path following task.}


\section{Problem Statement and Solution Overview}
\label{sec:problem}

The reference model for the walker kinematic is the unicycle,
described in discrete--time by the equations
\begin{equation}
\begin{cases}
    x(t_{k+1}) = x(t_{k}) + \cos(\theta(t_k)) \delta_t v(t_{k}), \\
    y(t_{k+1}) = y(t_{k}) + \sin(\theta(t_k)) \delta_t v(t_{k}), \\
    \theta(t_{k+1}) = \theta(t_{k}) + \delta_t \omega(t_k),
\end{cases}
\label{eq:model}
\end{equation}
where $\bm{q}(t_{k}) = \left[ x(t_{k}),y(t_{k}),\theta(t_{k}) \right]$
is the state of the vehicle, the coordinates $\left( x(t_{k}),y(t_{k})
\right)$ identify the position of the mid point of the rear wheel
inter-axle in the Cartesian plane $X_w \times Y_w$ expressed in the
$\frm{W} = \left\{ X_w,Y_w,Z_w \right\}$ world reference frame,
$\theta(t_{k})$ is the longitudinal direction of the vehicle with
respect to the $X_w$ axis, $v(t_{k})$ and $\omega(t_k)$ are the
longitudinal and angular velocities, respectively, and $\delta_t =
t_{k+1} - t_{k}$ is the sampling time. For the particular problem at
hand, $v(t_{k})$ is imposed by the human (also for safety
reasons~\cite{AndreettoDFFPZ18ral}), while $\omega(t_k)$ is the
control output and it is shared between the human and the robot. The
problem to solve is to control the vehicle from a starting position
$p_0 = [x_0, y_0]^T$ to a desired position $p_f$ in a known
environment. The key requirement is to use the robot controller
contribution to $\omega(t_k)$ only when the human behaviour deviates
significantly from the expected behaviour.

To this end, we need first to abstract the path following problem,
that is usually defined in the space $\bm{q}(t_{k})$, into a high
level representation that preserves the implicit features of the human
trajectories. Therefore, let us denote by
$\revv{\mathcal{H}}\subset\R^2$ the path travelled by the human in
$\frm{W}$, i.e., the sequence $\left( x(\revv{h}_{k}),y(\revv{h}_{k})
\right)$ of coordinates expressed with respect to the curvilinear
abscissa $\revv{h}$ sampled at times $\delta_t$. Let $\revv{\mathcal{R}}\subset\R^2$
be the reference path connecting $p_0$ to $p_f$. For both paths, we
extract a set of features of dimensionality $m$, denoted as
$\bm{z}_{k,\revv{\mathcal{H}}}$,~$\bm{z}_{k,\revv{\mathcal{R}}} \in
\mathbb{R}^{m}$, respectively, which are associated to a class of
human-like behaviours \revv{\{Left-turn, Right-turn, Straight\}}.

The overall framework of the proposed solution (sketched in
Fig.~\ref{fig:scheme}) comprises the following steps:
\begin{figure}[t!]
\centering
\includegraphics[width=\linewidth]{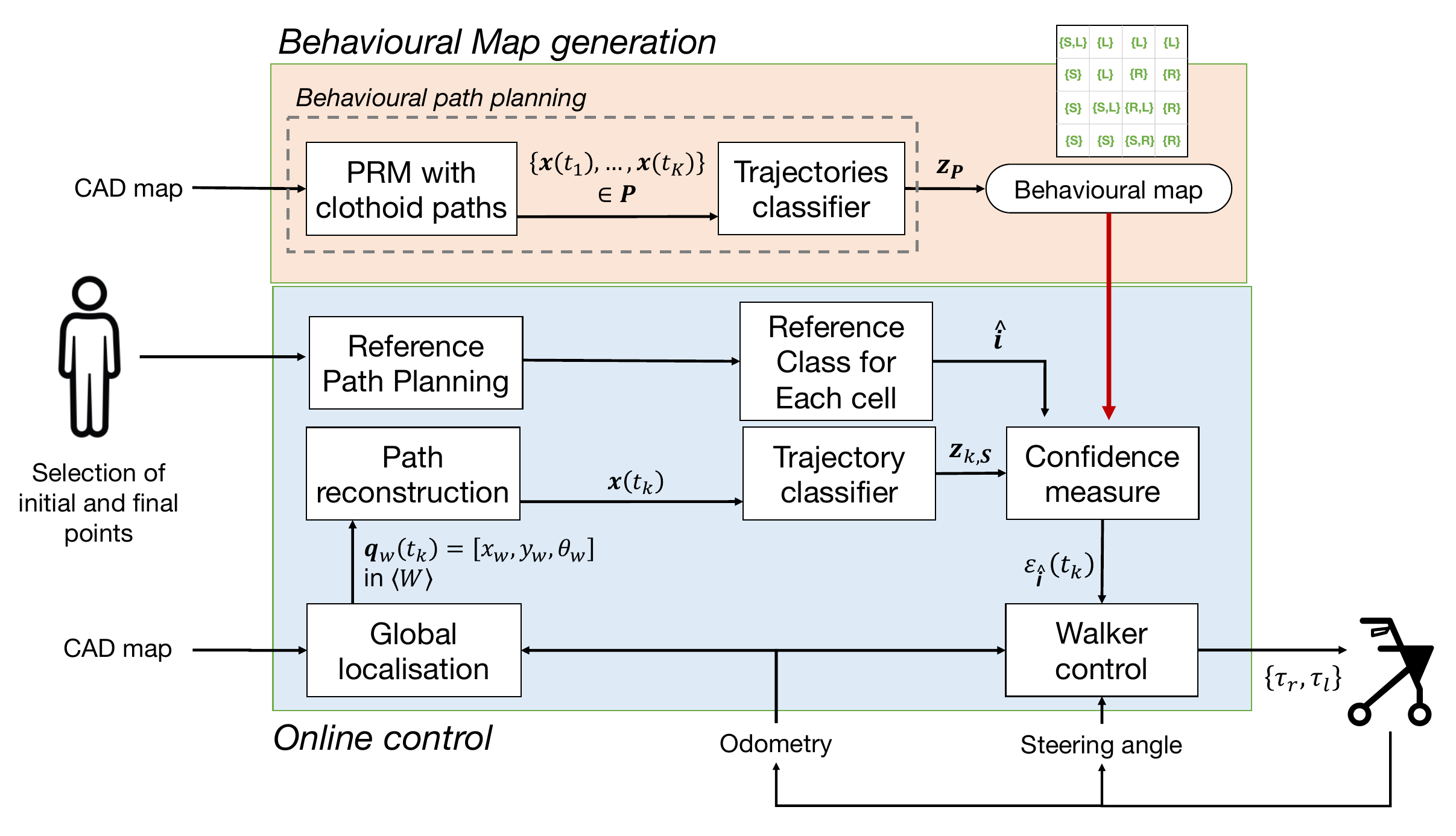}
\caption{Overall scheme of the algorithm.}
\label{fig:scheme}
\end{figure}

\revv{
\noindent
1. Given an a-priori map of the environment, composed of only
the static obstacles information (walls, furniture...), we generate a
large number of trajectories that are  collision free and that mimic a
human-like behaviour~\cite{Laumond_2010}. This operation is performed
offline and is showed in Fig.~\ref{fig:scheme} as the block: {\em PRM
with clothoid paths};

\noindent
2. We partition the map using square cells. Then we classify the
generated trajectories inside each cell into the \{Left-turn,
Right-turn, Straight\} classes by using a neural network ({\em
Trajectories classifier} block in Fig.~\ref{fig:scheme}). This
classification is used to create a map of possible human-like
manoeuvres that can be used to reach a specific location. This grid map
will be called the {\em behavioural map} ({\em Behavioural map} block
in Fig.~\ref{fig:scheme}).

\noindent
3.  Given the user initial position $p_0$, and a user--selected final
position $p_f$ we connect them with a trajectory (the planning method
will be explained later in the paper). This trajectory is
characterised by its features $\bm{z}_{k,\revv{\mathcal{H}}}$. For
each portion of the ideal trajectory, we identify a reference cluster
of trajectories generated at Step 1 and 2. From this cluster, we
select a set of representative features
$\bm{z}_{k,\revv{\mathcal{R}}}$ (e.g., the centroid of the cluster).
This allows us to compute the likelihood $\varepsilon(t_{k})$ ({\em
Confidence measure} block in Fig.~\ref{fig:scheme}) and use it to
generate the setting of the visco-elastic force used in our
shared-authority control scheme ({\em Walker control} block in
Fig.~\ref{fig:scheme}).
}


\section{Model generation and behaviour--based control}
\label{sec:model}
The main pillars of our approach are an offline analysis of the
environment that generates the behavioural map (i.e., the map of
admissible behaviours for every area of the environment) and the
online control module that adapts the shared authority controller to
the degree of compliance of the user. The two modules are described
next.

\subsection{Behavioural map generation}
\label{subsec:map-generation}

Given the environment map, the behavioural map associates each area of
the space with the class of trajectories \revv{(straight, right turn, left
turn)} possibly followed by humans when they behave ``correctly''. This
information is generated in different steps.

In the first step, we generate a Probabilistic Road Map
(PRM)~\cite{LatombeKSOtra} covering the entire space. The PRM provides
collision free geometric paths connecting any pair of locations in the
space.  The PRM is generated ensuring an average density of 4
nodes per squared meter, which is a good trade-off between fine
distribution of nodes and elaboration time of the paths (e.g.  in a
5x5 meters room we have an average of 100 nodes). 

In the second step, we consider pairs of random starting positions and ending
positions, find the shortest path connecting them through the PRM, and
interpolate the different nodes by clothoids. A \emph{clothoid} is a
line with curvature proportional to the arc-length described by the
equation
$X(s) = x_0\! + \int_0^s \cos ( \kappa' \frac{\tau^2}{2}+\kappa_0 \tau
+ \theta_0) \textrm{d}\tau$, $Y(s) = y_0 + \int_0^s \sin ( \kappa'
\frac{\tau^2}{2}+\kappa_0 \tau + \theta_0) \textrm{d}\tau$,
where $s$ is the curvilinear abscissa, $(x_0, y_0)$ is the Cartesian
coordinate of the initial point, $\theta_0$ is the initial bearing,
$\kappa_0$ is the initial curvature and $\kappa'$ is the change rate
of the curvature. The interpolation is done minimising the derivative
of the squared curvature~\cite{bertolazzi2018g2}.  We can argue that
the trajectories constructed in this way are a reasonable
approximation of human-like trajectories, as supported by numerous
results in the literature. The most important are in the work of
Laumond et al.~\cite{Laumond_2010}, in which clothoids are explicitly
addressed as a good approximation of human trajectories, and in the
work of Arechevaleta et al.~\cite{arechavaleta2008optimality}, who
have shown that humans tend to minimise the derivative of the squared
curvature when they move.

\begin{figure}[t!]
\centering
\includegraphics[width=\linewidth]{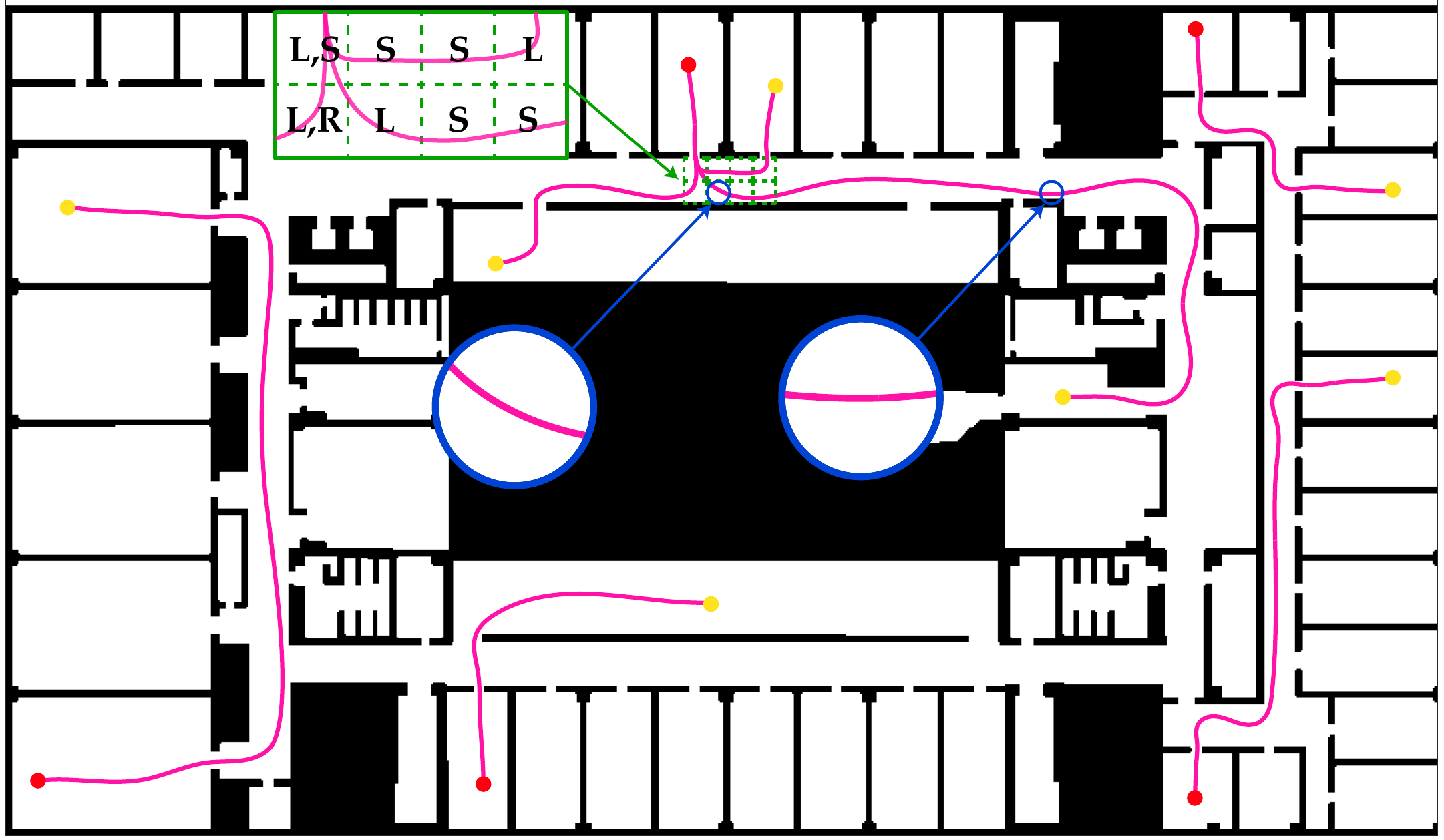}
\caption{Synthetic trajectories generated with PRM and clothoids.}
\label{fig:trajs}
\end{figure}

In the third step, the environment is discretised in a grid map using
1x1 meters cells. For each trajectory $i$ intersecting a cell $j$, we
identify a class $c_i^{(j)}$ in the finite set $\bm{c}$:
$c_i^{(j)}\in\bm{c}$. For instance, one class could be ``left turn''
(L) or ``move straight'' (S).  This operation is performed by the {\em
  Trajectory Classifier}, which allows us to partition each trajectory
into a sequence of elementary moves (straight, left/right turn)
\revv{and determine the class that identify each of them in every
  cell.} To account for the different direction of motion of the
$i$-th trajectory within the $j$-th cell, we associate the tangential
direction $\theta_i^{(j)}$ \revv{, which is the mean direction of
  travel of the vehicle in the $j$-th cell w.r.t the map reference
  frame,} with the class $c_i^{(j)}$. \revv{For instance if the user
  is moving with the walker straight from west to east the tangential
  direction $\theta_i^{(j)}$ will be 0.} The set of all the pairs
$\left(c_i^{(j)}, \theta_i^{(j)}\right)$ form the behavioural map. In
Figure~\ref{fig:trajs} some of the synthetic trajectories are shown in
magenta in a map of the Department of Engineering and Computer Science
of the University of Trento, the grid map is shown in green and the
corresponding sub-trajectories that will be fed to the classifier are
highlighted with the blue circles. \revv{The starting points of the
  magenta trajectories are depicted with the red circles, while the
  ending points with the yellow circles.}
The fundamental motion primitives forming the behaviour map are shown
in light green in Figure~\ref{fig:trajs}.

\noindent {\bf The trajectory classifier. } The trajectory features
are extracted with an encoder neural network from the path geometry.
More precisely, the $k$-th abscissa $s_{k}$ of the path $\revv{\mathcal{R}}$,
sampled such that $s_{k} - s_{k-1} = \delta_s$ is constant, is used to
define the vector of geometric parameters
\begin{equation}
  \bm{p}(s_{k}) =
    \begin{bmatrix}
     x_p(s_{k})\\
     y_p(s_{k})\\
     \cos(\theta_p(s_{k}))\\
     \sin(\theta_p(s_{k}))\\
     \kappa_p(s_{k})
  \end{bmatrix},
  \end{equation}
where $x_p(s_{k})$ and $y_p(s_{k})$ are the Cartesian coordinates,
while $\theta_p(s_{k})$ and
$\kappa_p(s_{k})={d\theta_p(s_{k})}/{ds_{k}}$ are the tangential axis
and the curvature of $\revv{\mathcal{R}}$ in $\left( x_p(s_{k}), y_p(s_{k})
\right)$, respectively.  To account for the path characteristics, $n$
consecutive parameters are collected on the sampled abscissa
coordinates $s_{k-(n-1)}$ to $s_{k}$, so as to build the matrix
comprising $\bm{p}(s_{k-(n-1)})$ to $\bm{p}(s_{k})$, which is then
normalised to avoid spatial biases, i.e. 
\begin{equation}
\bm{x}_p(t_{k}) =
  \begin{bmatrix}
  \left[x_p^{1}, \dots, x_p^{n}\right] - x_p^{1}{\textbf 1}^{T} \\
  \left[y_p^{1}, \dots, y_p^{n}\right] - y_p^{1}{\textbf 1}^{T} \\
  \cos(\theta_p^{1}), \dots, \cos(\theta_p^{n}) \\
  \sin(\theta_p^{1}), \dots, \sin(\theta_p^{n}) \\
  \kappa_p^{1}, \dots, \kappa_p^{n} 
\end{bmatrix},
\label{eq:nn-inputs}
\end{equation}
where ${\textbf 1}$ is an $1$-dimensional column vector with all ones,
used for the normalisation of the position vectors, and we adopt the
compact notation $x_p^{i} = x_p(s_{k-(n-i)})$. In order to avoid the
problem of angular periodicity, we used both
$\cos(\theta_p(s_{k-(n-i)}))$ and $\sin(\theta_p(s_{k-(n-i)}))$
instead of $\theta_p^i$.

In the training process of the encoder, $\bm{x}_p(t_{k}) \in
\mathbb{R}^{5 \times n}$ is used as input. The weights of the encoder
are learned by training an autoencoder and minimising the
reconstruction error between $\bm{x}_p(s_k)$ and the reconstructed
output $\tilde{\bm{x}}_p(s_k)$. The encoder and the decoder
sub-networks of the autoencoder, have a symmetrical structure: the
input $\bm{x}_p(t_{k})$ passes through $3$ convolutions and $3$
fully-connected layers, resulting in a final latent space of $m=5$
neurons. The decoder, then, has the same structure, but takes as input
the latent space $\bm{z}_{k,\revv{\mathcal{R}}} \in \mathbb{R}^{5}$.

After learning the autoencoder weights, the decoder sub-network is
discarded as we will use the latent space of the autoencoder as a
compressed representation of the human behaviour (\emph{Net1}). A second neural
network (\emph{Net2}) classifies $\bm{z}_{k,\revv{\mathcal{R}}}$ into the behavioural
classes in the set $\bm{c}$. More precisely, the behaviour is
identified in the minimalistic set \{Left-turn, Right-turn, Straight\}
and encoded by the numeric label $\bm{c} = \left\{1,2,3\right\}$.
Hence, during the learning phase we define the transformation
\begin{equation}
    \tilde{\bm{c}} = h_{\psi} \left( \bm{z}_{k,\revv{\mathcal{R}}} \right) ,
    \label{eq:classifier}
\end{equation}
\revv{where $\psi$ is a set of parameters obtained by minimising the
cross--entropy} between the predicted $\tilde{\bm{c}}$ and the actual
$\bm{c}$ class. In the architecture of the classifier network the
input latent feature $\bm{z}_{k,\revv{\mathcal{R}}}$ of $5$ neurons passes
through a single fully-connected layer with just $3$ neurons.
Therefore, the combination of the neural networks maps the geometric
characteristics of the path $\bm{x}_p(s_k)$ into the trajectory
classes encoded in $\bm{c}$. As a final step, a $\softmax(\cdot)$
activation function is \revv{applied to the three output neurons of
\emph{Net2} } to retrieve the confidence $\varepsilon_i(s_{k})$ (or
equivalently $\varepsilon_i(t_{k})$) of the class $c_i\in \bm{c}$,
i.e. $\sum_{i=1}^3\varepsilon_i(s_{k}) = 1$. Notice that this same
network is adopted to classify the synthetic generated paths and the
current user behaviour, as will be explained in the next
Section~\ref{subsec:control}.

\revv{
\paragraph{Geometry of the Grid}
  In the discussion above we suggest a decomposition into a grid made
  of square cells. This choice is not mandatory. For other types of
  environments with the presence of static obstacles with
  non-rectangular shape, it could be more convenient to use a
  different type of cell decomposition (e.g., maximum clearance maps,
  maps resulting from plane sweep, etc.)~\cite{lavalle2006planning}.
  The technique proposed in the paper would not be significantly
  affected by the choices of a different polygonal geometry.
}

\subsection{Online Control}
\label{subsec:control}
As a first step in the online control, the user selects their
destination ($p_f$) starting from the initial point $p_0$ where the
device is currently located. 
\revv{It is prudent to impose limits on the maximum traveled distance
when dealing with individuals who may have limited mobility or other
frailties. Equally important is the optimisation of routes connecting
subgoals based on specific metrics. We've previously tackled these
challenges and provided solutions in our earlier
work~\cite{bevilacqua2020activity}. In our current research, we assume
that all pertinent decisions regarding these constraints and
optimisation criteria have been predetermined before executing our
algorithm.}

Following the same steps as for the
behaviour map generation, the system connects the two points
\emph{via} the PRM and interpolates the intermediate points by using a
G2 spline that minimises the derivative of the squared curvature. This
allows us to determine for the current cell $j$ the reference class
and its orientation $(c^{(j)}_{\hat{i}}, \theta^{(j)}_{\hat{i}})$.
Roughly speaking, this pair encodes the most sensible behaviour that a
human would follow if they want to reach $p_f$ from the cell $j$, and
will be used to measure the degree of compliance of the human.

A custom path reconstruction module, described
in~\cite{antonucci2021humans}, processes the odometry information
received by the {\em FriWalk} and produces in real--time \revv{$\mathcal{H}$}
and, hence, the sets $\bm{x}_s(s_{k})$. The currently performed path
$\bm{x}_s(s_{k})$ is reduced to its features $\bm{z}_{k,\revv{\mathcal{H}}}$
using the neural network explained in the previous
Section~\ref{subsec:map-generation}.  These features are compared to
the ones stored in the {\em Behavioural map} to obtain the confidence
value $\varepsilon_{\hat{i}}(s_k)$ ({\em Confidence measure} in
Figure~\ref{fig:scheme}). A high confidence means that the features of
the user motion are compatible to a large extent with the class
$c^{(j)}_{\hat{i}}$, while a low confidence means that the user is drifting away
from the expected behaviour.  Hence, $\varepsilon_{\hat{i}}(s_k)$ will be
used as a hyper-parameter in the control of the Walker.

The control module is designed synthesising a visco--elastic torque
that is applied to the steering angle of the front wheels of the
robot.  \revv{The idea of the visco-elastic
control~\cite{andreetto2019authority} can be described as follows.
Suppose that the path synthesised by the system (desired trajectory)
is described by means of the desired steering angle $\theta^\star$ and
steering velocity $\dot{\theta}$. Let $\theta$ and $\dot{\theta}$ be
the actual measured or estimated values. The torque applied to the
system is given by
\[
  \tau = - a\left(\theta-\theta^\star\right) - b\left(\dot{\theta}-\dot{\theta}^\star\right) .
\]
In simple words, the vehicle is governed by a torque generated by a
spring-damper system. Importantly, the spring constant $a$ and the
damper constant $b$ are not time invariant but are functions. In our
original idea~\cite{andreetto2019authority} these functions depend on
the deviation from the desired path: the larger the deviation from the
desired path the stiffer become the controller. This controller is
practically stable, meaning that it secures the convergence of the
error to a neighbourhood of the origin.}

To this end, we first define the actual right (left) wheel
angle as $\alpha_{r}$ ($\alpha_{l}$), which can be measured by an
absolute encoder (we dropped the reference to the time $t_k$ for ease
of notation). The states are expressed in the robot reference frame
$\frm{R} = \left\{ X_r,Y_r,Z_r \right\}$, with $X_R$ oriented along
the longitudinal direction and $Z_R$ pointing upwards. The desired
wheel angles $\alpha_{r}^\star$ and $\alpha_{l}^\star$, instead, can
be obtained by the desired angular velocity $\omega^\star$ (computed
using the behaviour map associated with the desired class
$(c^{(j)}_{\hat{i}}, \theta^{(j)}_{\hat{i}})$), the actual
longitudinal velocity $v$ (obtained by the encoders on the rear
wheels) and the Ackermann steering geometry. Hence, the wheels
orientation errors
\[
  e_{\alpha_r} = \alpha_{r}^\star - \alpha_{r} \mbox{ and }
  e_{\alpha_l} = \alpha_{l}^\star - \alpha_{l} ,
\]
can be immediately obtained. The visco-elastic controller that
controls the torque to apply to the wheel is determined as
\begin{equation}
  \label{eq:TorqueRight}
  \tau_{\alpha_r} = a e_{\alpha_r} + b \dot{e}_{\alpha_r} ,
\end{equation}
and the same for the left wheel to obtain $\tau_{\alpha_l}$.

While this controller accounts for the local direction that the
vehicle has to take w.r.t. $\frm{R}$ (i.e., the curvature of the
path), we also compute the absolute orientation error of the front
wheels in $\frm{W}$. Again, using the desired orientation
$\theta_{\hat{i}}^{(j)}$, and thus compute the desired wheel direction
$\beta_r$ in $\frm{W}$ as
\[
  e_{\beta_r} = (\theta^\star + \alpha_{r}^\star) - (\theta +
  \alpha_{r}) .
\]
The same can be applied to $\beta_l$. Hence, we can compute the same
visco-elastic controller in~\eqref{eq:TorqueRight} for the errors
$\beta_r$ and $\beta_l$ (with the same parameters $a$ and $b$), thus
obtaining the final control laws
\begin{equation}
  \label{eq:FinalTorques}
  \tau_{r} = \lambda\tau_{\alpha_r} + \tau_{\beta_r} \mbox{ and } \tau_{l} =
  \lambda\tau_{\alpha_l} + \tau_{\beta_l} .
\end{equation}

The parameters $\lambda$, $a$ and $b$ are functions of the confidence
$\varepsilon_{\hat{i}}(t_k)$ associated with the reference class
$c_{\hat{i}}^{(j)}$ by means of
\begin{equation}
  \label{eq:ContLawParams}
\begin{aligned}
  \lambda = 1 - \varepsilon_{\hat{i}}(t_{k}), \\
  a = a_0 + a_1 \lambda, \\
  b = b_0 + b_1 \lambda .
\end{aligned}
\end{equation}
\revv{The $a$ parameters influence the elasticity of the control
law, while the $b$ parameters influence its viscosity. Specifically,
$a_0$ and $b_0$ are the minimum coefficients used when the controller
does not intervene  (when the confidence is high). The $a_1$ and $b_1$
are modulated by the hyperparameter epsilon (the confidence). This
means that increasing values of higher $a_0$ and $b_0$ will result in
more intervention from the control, even when the human is performing
correct movements. Similarly, the values of $a_1$ and $b_1$ widen or
shrink the range of the applied control signal between when the system
intervenes and when it does not. These parameters have to be
fine-tuned by trail-and-error sessions on the specific application.}

To summarise, we first compute the reference class $c^{(j)}_{\hat{i}}$, then
from the actual state $\bm{x}_s(t_{k})$ we compute
$\varepsilon_{\hat{i}}(t_{k})$ and then, by means
of~\eqref{eq:ContLawParams}, the desired torques are computed
with~\eqref{eq:FinalTorques}.  The term $b_0$ is needed to avoid
oscillatory behaviours while $a_0$ is needed to generate the correct
control signal that forces the wheel angle $\alpha_r$ ($\alpha_l$) to
the desired value. In this way, when the confidence is high (i.e.,
$\lambda$ is low), the applied torque is predominantly imposed by the
user and the computed torques $\tau_{r}$ and $\tau_{l}$ tend to zero.
The system, instead, becomes increasingly authoritative (i.e., torques
$\tau_{r}$ and $\tau_{l}$ imposed by the system) when $\lambda$ gets
closer to \revv{$1$}.

\revv{
\paragraph{Management of obstacles and of exceptions}
The approach outlined above hinges on the definition of a reference
trajectory for each cell (given by the centroid of the most probable
cluster) and on the application of visco-elastic control to make sure
that the user does not deviate too much.  Two type of exceptions can
occur: 

\noindent
1. An unexpected dynamic obstacle (e.g., another human) materialises, 

\noindent
2. The user strongly opposes the suggestion and forces her/his way.  

\noindent
The first case is handled by using the so called reactive
planning~\cite{BevilacquaFFP18ral}: the system replans a new
clothoidal trajectory that travels around the obstacle and joins into
the reference trajectory as soon as the obstacle is overcome. This
change has no significant impact on the framework: we can either use
the visco-elastic control modulated by the likelihood $\epsilon$ or
opt for a stiffer behaviour until the anomaly is over.  For the second
exception, we interpret the strong opposition of the user as her/his
better understanding of the scenario. Therefore, we disengage the
guidance system for a reconfigurable time. This choice does not apply
if the user is travelling across areas that we deem dangerous (e.g., a
stairway).
}


\section{Experimental Validation and Results}
\label{sec:experiments}

\noindent {\bf Generation of the behavioural map.}  The experimental
validation of the approach has been carried out in our Department
premises. The first step of the approach was the construction of the
behaviour map generating the described human-like synthetic
trajectories (some examples are shown in Figure~\ref{fig:trajs}). For
the training of \emph{Net1} and \emph{Net2} we focused on an area
consisting of two intersecting corridors (conventional
cross--intersection).
We simulated $1800$ paths selecting randomly pairs of waypoint
positions $p_0$ and $p_f$. The simulations were equally partitioned in
the Left-turn, Right-turn and Straight classes. We select $n=12$
samples for the inputs $\bf{x}$ \revv{, as shown in Eq. (3), this
implies that the encoder will have 60 input features}. A step size
of $\delta_s=0.1$~m has been chosen to sample the trajectory.
A fraction of $80\%$ of the dataset was used as training set, while
the remaining samples were randomly selected for the validation. Both
{\em Net1} and {\em Net2} were implemented in Keras and trained with
the Adam optimiser with a learning rate of $0.001$, batch size $64$,
and number of epochs $300$ using a 2.7 GHz Intel Core i7 processor.
{\em Net1} was trained using the set of $\bf{x}$ as both inputs and
outputs of the network. Then, we transferred the learned weights of
the encoder in the {\em Net2}, and performed a supervised training by
comparing its estimates with the \revv{one-hot} encoded labels of
classes \revv{\{Left-turn, Right-turn, Straight\}.}

In Table~\ref{table:training-unsupervised}, we report the inference
accuracy of the network {\em Net1} on the validation set, in terms of
Root Mean Squared Error (RMSE).
\begin{table}[t]
    \centering
    \caption{RMSE of the {\em Net1} and the precision discovery rate
      of the {\em Net2} on the validation set of the synthetic
      trajectories.}
    \label{table:training-unsupervised}
    \begin{tabularx}{\linewidth}{cccccc}
    \toprule
    & $x$~(m) & $y$~(m) & $\cos(\theta)$ & $\sin(\theta)$ & $\kappa$
    \\
    \midrule
    RMSE & 0.0076 & 0.0118 & 0.0293 & 0.0449 & 0.0241 \\
    \midrule
    & Left & Right & Straight & Average & \\
    \midrule
    \thead{\revv{Accuracy}} & \revv{88.4}$\%$ & \revv{88.3}$\%$ & \revv{76.8}$\%$ & \revv{84.3}$\%$ & \\
    \bottomrule
    \end{tabularx}
\end{table}
The results show that the network was correctly trained on the
dataset, and the even distribution of the error over the different
components of the input indicates that no bias was produced in favour
of a particular component. The parameters $a$ and $b$, as explained in
Section~\ref{subsec:control}, are functions of the confidence of the
manoeuvre. For the angular velocity $\omega$, the parameters of the
visco-elastic controller in~\eqref{eq:ContLawParams} were set to
$a_0=25$~N, $a_1=15$~N, $b_0=15$~Ns and $b_1=10$~Ns. For the steering
wheel direction $\beta$, the parameters of the visco-elastic
controller were set to $a_0=25$~N, $b_0=25$~Ns.  These parameters were
set leveraging our experience with the system.  Changing such
parameters, modifies the amount of intervention of the robot control.

The validation results for the training of the network {\em Net2} are
reported in Table~\ref{table:training-unsupervised}, showing the
accuracy of the inferred classes. It can be noticed that the Left and
Right classes obtained a higher percentage with respect to the
Straight class: the reason behind this behaviour is that the
trajectories of the Straight class include features in common with the
ones of the other classes (e.g., when the human slightly bends along
an almost straight path). This is noticeable in
Figure~\ref{fig:trajs}, where the sub-trajectories not always are
distinguishable between turns and straight sectors.

\subsection{Experiments with the {\em FriWalk}}
\label{subsec:experiments-robot}

\begin{figure}[t]
	\centering
	\begin{tabular}{ m{0.3cm} m{5.5cm} } (a) & \hspace{10pt}
   \includegraphics[width=0.8\linewidth]{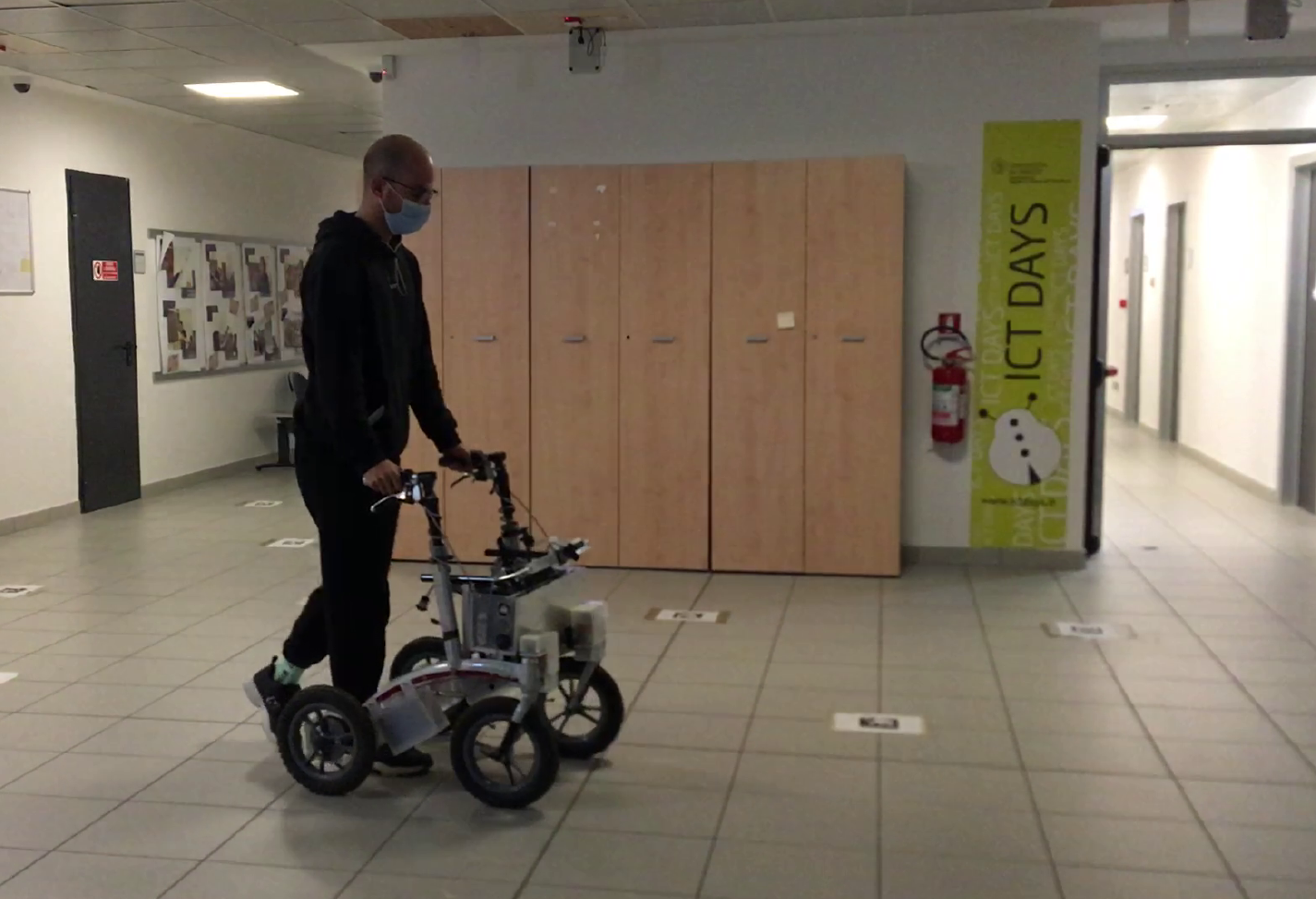} \\
   (b) &
   \includegraphics[width=0.9\linewidth]{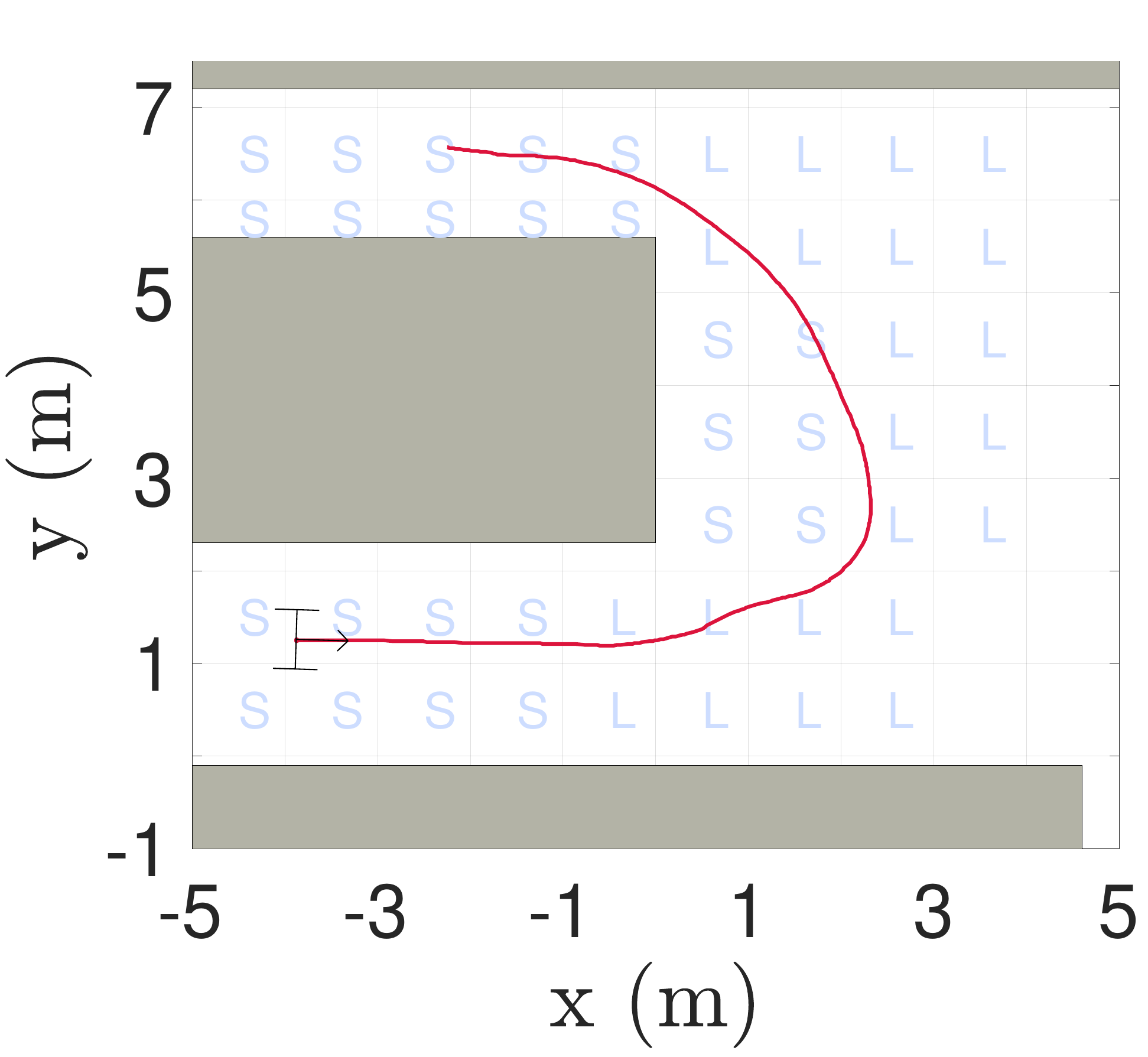} \\
	\end{tabular}
	\caption{(a) Photo of the experimental area and (b) the associated behavioural map.}
	\label{fig:experiments-walker-1a}
\end{figure}

\begin{figure}[t]
	\centering
	\begin{tabular}{ m{0.1cm} m{8cm} } (a) &
		\includegraphics[width=0.8\linewidth]{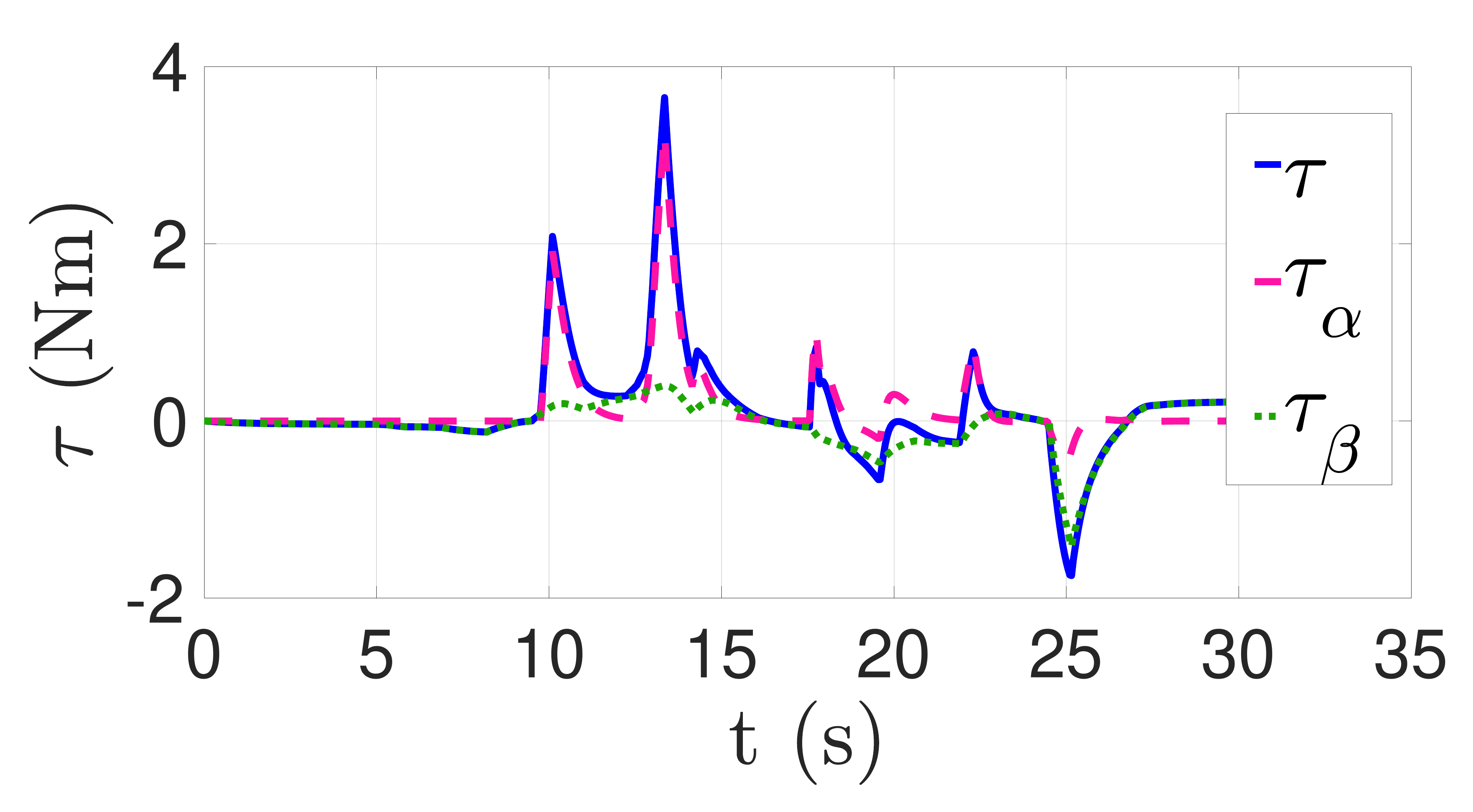}
		\\
		(b) &
		\includegraphics[width=0.8\linewidth]{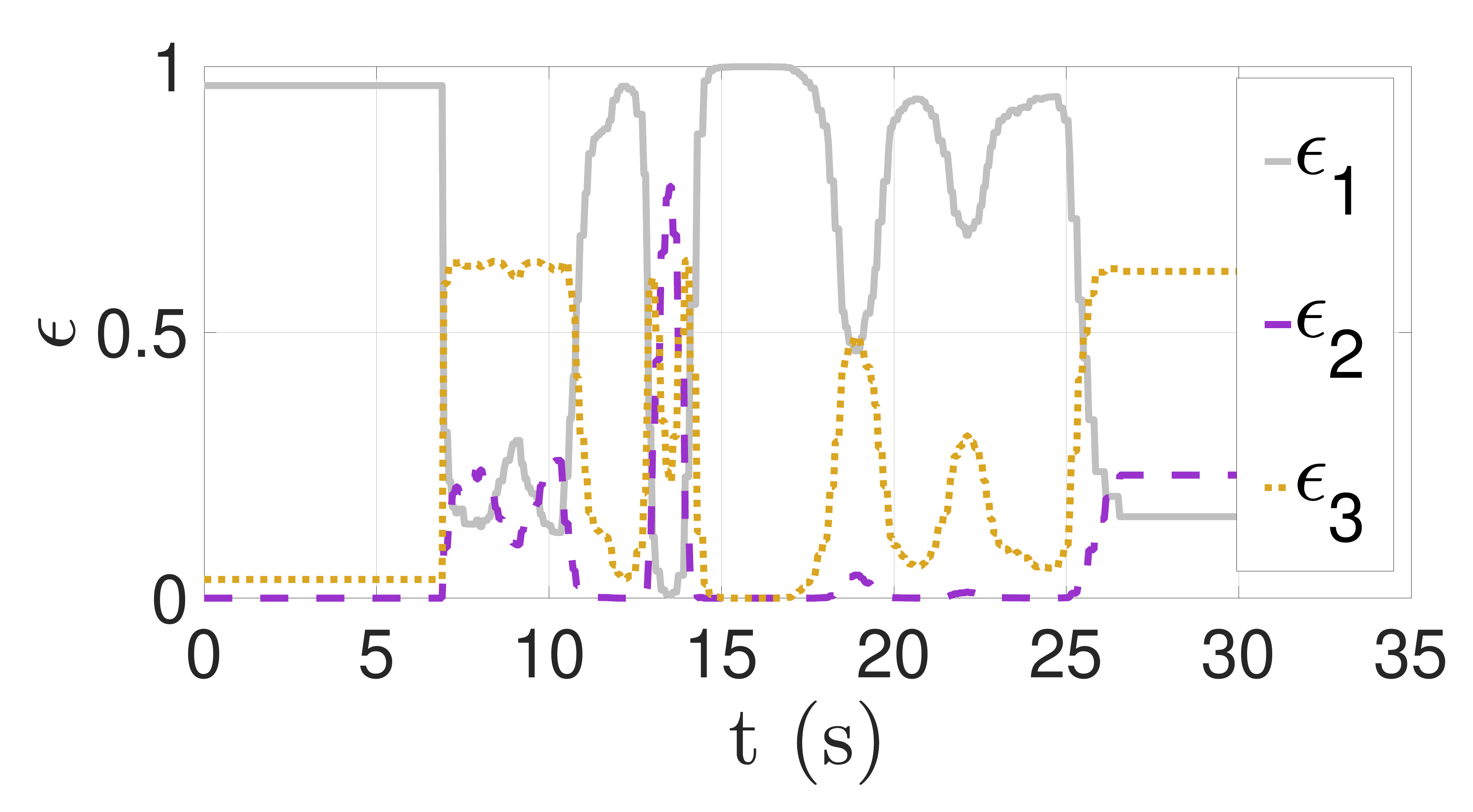}
		\\
	\end{tabular}
	\caption{(a) Torque controls for $e_{\alpha_r}$ and $e_{\alpha_l}$
		(magenta-dashed line) and for $e_{\beta_r}$ and $e_{\beta_l}$
		(green-dotted line) applied to the walker front wheels while
		performing the trajectory in
		(Fig~\ref{fig:experiments-walker-1a}-b). (b) confidence for
		Left-turn (grey-solid line), Right-turn (purple-dashed line) and
		Straight (yellow-dotted line).}
	\label{fig:experiments-walker-1b}
\end{figure}

The experimental evaluation of the approach presented in
Section~\ref{sec:problem} and Section~\ref{sec:model} was conducted on
the real {\em FriWalk} in an indoor hallway at the University of
Trento. The vehicle is endowed with front electric DC motors to
control the angle of the front wheels.  The localisation system of the
robot comprises incremental encoders in the rear wheels and absolute
encoders for the front wheels, used in combination with a 2D camera
system. A collection of ArUco markers was placed in the testing area,
which has a dimension of roughly $7\times7$~m, allowing the walker to
localise itself with sufficient accuracy (error below $20$~cm, as
reported in~\cite{NazemzadehFMP17tmech}). A ROS interface was used to
send the control to the actuators and to receive the localisation
data, including the odometry-based estimates of $\bm{q}(t_{k})$ in
$\frm{W}$ and the angular position of the wheels $\alpha_r$ and
$\alpha_l$ in $\frm{R}$.  We fixed the initial and final waypoint
areas for the tests and we executed offline the behavioural path
planning described in Section~\ref{subsec:map-generation}, obtaining
the mentioned behavioural map. We then executed several trials of the
same mission, varying the general behaviour of the human experimenter
between three macro categories: following diligently the predefined
mission, following the mission roughly and deviating from the mission.
Figure~\ref{fig:experiments-walker-1b}-a shows the control action of
the robot while the human moved for the leftmost corridor towards an
exit on the upper part of the map (see
Figure~\ref{fig:experiments-walker-1a}-b).
After $10$~seconds from the beginning of the experiment, the user kept
walking straight an area where the Left-turn class was instead
foreseen: the low likelihood on the Left behaviour (grey line in
Figure~\ref{fig:experiments-walker-1b}-b) triggered a compensating
action on the control signal $\tau_{\alpha_r}$ and $\tau_{\alpha_l}$
in~\eqref{eq:FinalTorques} (dashed magenta curve in
Figure~\ref{fig:experiments-walker-1b}-a), thus causing a compensation
in the trajectory. This intervention results into the increasing
likelihood of the Left-turn behaviour class (grey line in
Figure~\ref{fig:experiments-walker-1b}-b). Similarly, as the human
tried to steer right after $13$~seconds, the corresponding Right-turn
behaviour was caught (purple dashed line in
Fig.~\ref{fig:experiments-walker-1b}-b) and the authority was again
transferred to the robot, i.e., the human was progressively pushed
towards the correct turning behaviour. Notice that when the
compensation action occurs, the human user corrects the erratic
behaviour in a few instants, indulging the robot action and indirectly
lowering the control action. Hence, the robot action is perceived as a
brief suggestion that vanishes immediately if the user follows the
change of the route, otherwise the control action will persistently
assist the manoeuvre towards the correct direction.

\begin{figure}[t!]
  \centering
	\begin{tabular}{ m{0.1cm} m{8cm} } (a) & \hspace{15pt}
      \includegraphics[width=0.6\linewidth]{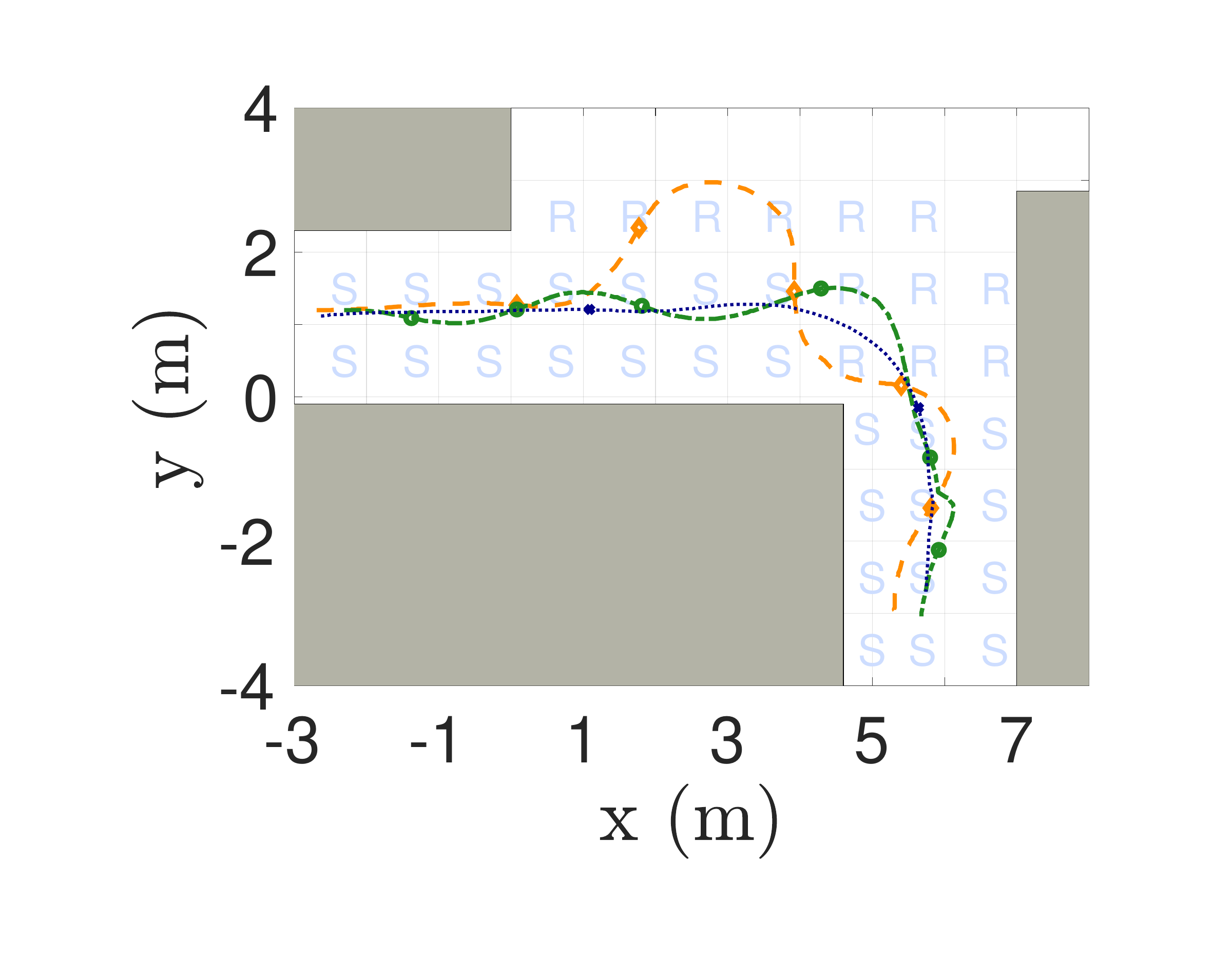} \\
    (b) &
    \includegraphics[width=0.8\linewidth]{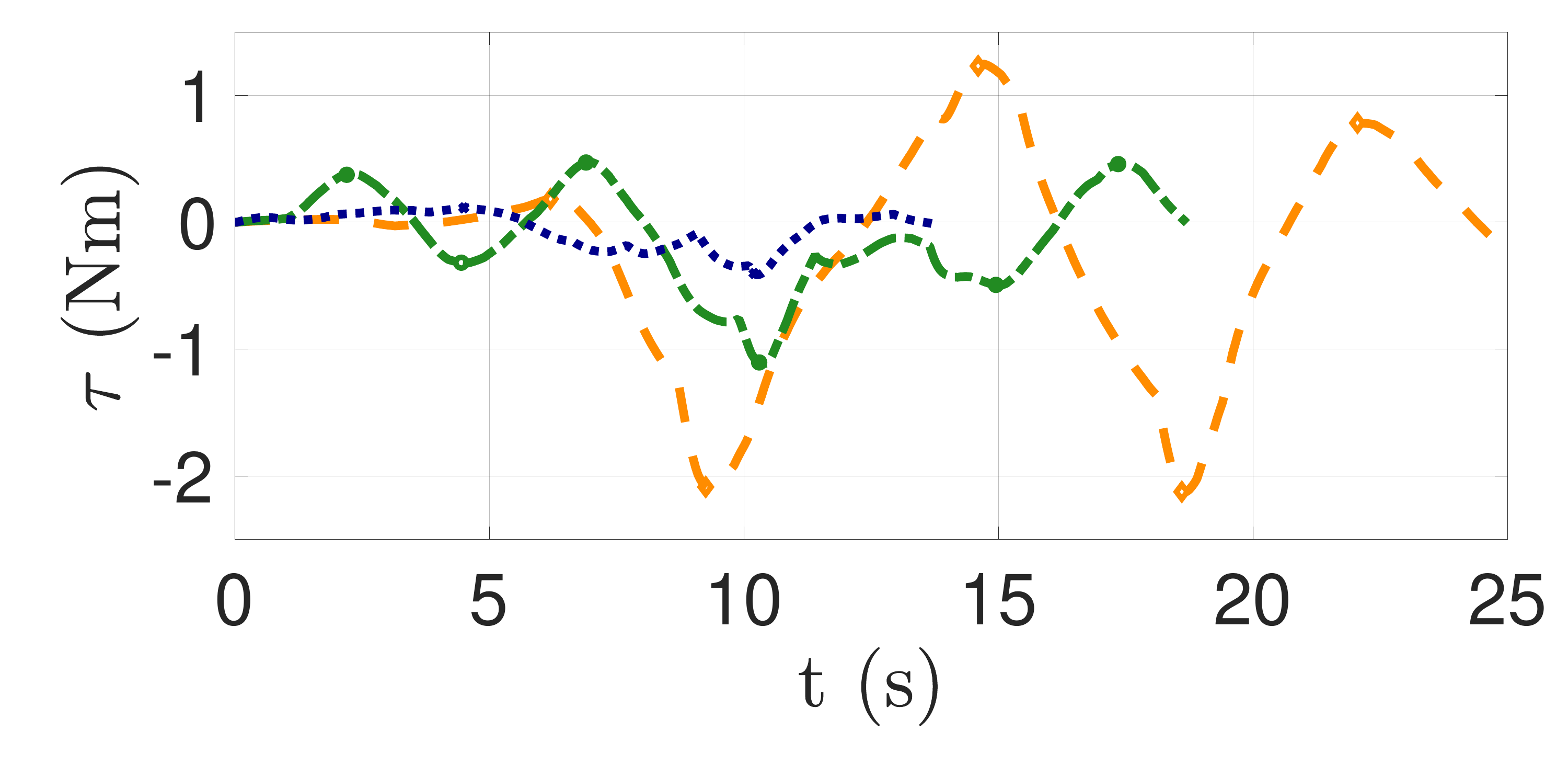} \\
  \end{tabular}
  \caption{Experimental evidence of the control action behaviour in
    case of a user acting purposely against the desired path (orange
    lines), making slight deviations (green lines) or adhere to the
    planned path (blue lines). The resulting path (a) and the relative
    control actions (b) are reported.}
  \label{fig:experiments-walker-2}
\end{figure}

In Figure~\ref{fig:experiments-walker-2}, we depict the performance of
the control for three different user's behaviours.

When the person is compliant with the planning (blue trajectory), the
control does not intervene, so the person is fully in charge and do
not feel any opposing action from the robot. When, instead, the user
purposely acts against the planned path, the control actions are
extremely evident (orange trajectory).  Finally, in the most typical
case, the control acts loosely without excessively forcing the path
correction (green trajectory), keeping the motion in the appropriate
direction (please also refer to the multimedia complementary material
accompanying this paper for further examples).

\subsection{User evaluation}
\label{subsec:user-evaluation}

Since this work hinges for a large part on human-robot interaction, a
qualitative evaluation of the system behaviour was needed to validate
the user acceptability. Simple experiments were defined to propose the
control strategy to participants, which were followed by a brief poll
to evaluate the level of appreciation of the person. All the
participants were informed that data collection and the information
provided are covered by the ethical rules of the Research Ethics
Committee, which approved this experiment, and that they could quit
the experiment at anytime. Once consent was obtained they were invited
to perform the tasks with the FriWalk. The experiment was performed by
16 adults \revv{in the age range of 21-50 years old,} without motor
nor cognitive impairments. \revv{The participants were 11 males and 5
females, all from the University of Trento. The participants do not
use walking aid devices in their daily lives.}

The experiments were held in a single experimental session for each
participant with a maximum duration of $15$ minutes, divided into
three main parts. The participant tried the walker with two navigation
techniques: the solution here presented and a visco--elastic control
applied to force the vehicle on an optimally planned
path~\cite{andreetto2019authority} used as comparison. Both navigation
techniques were applied to the same indoor environment, with the same
points of departure and arrival and both had similar performance and
efficiency in terms of data requirements, computing power, and
scalability.  Moreover, the order of presentation of the two
techniques was randomised to obtain comparable results from the polls,
and to the participants they were presented as navigation technique A
and navigation technique B. 
\revv{The participant would not initially be told which of the two
navigation techniques is the result of this study so as to avoid
influencing the perceptions of the driving experience. The
participants were asked to walk naturally with the aid of the
walker from their current initial position to a specific point
showed to them. They were asked to move compliantly to the target
and then a in a second attempt to move erratically to the target
or even go to the wrong direction.}
Finally, the participant was asked to answer a short questionnaire in
which a qualitative evaluation of the aspects of the two navigation
methods tried, and a final question were asked, in which these methods
were compared. The participant could decide to repeat the navigation
tests several times before the polls if it was necessary to achieve a
better understanding of its functioning, \revv{but none of them asked
to repeat the test}. The participant could move freely in the
environment, keeping in mind that the navigation techniques under test
have no mechanism to avoid obstacles.  It was the participant who took
charge of avoiding hitting the walls and any other obstacles that may
arise during the experiment.

In Table~\ref{table:user-evaluation1} and
Table~\ref{table:user-evaluation2} are reported the poll results.
\begin{table}[t]
\caption{User evaluation (yes)}
\begin{tabular}{ccc}
\toprule
    Question & \thead{Visco--elastic\\ control} & \thead{Behavioural\\
    maps control}  \\
    \midrule
    \makecell{Was it evident\\ that was the walker\\ to decide the
    path\\ to follow?} &  $87.5\%$ & $12.5\%$  \\
    \midrule
     \makecell{Have you felt\\ to be pulled,\\ pushed or stuck?} &
     $37.5\%$ & $12.5\%$  \\
    \midrule
\end{tabular}
\label{table:user-evaluation1}
\end{table}
\begin{table}[t]
\caption{User evaluation (mean - standard deviation)}
\begin{tabular}{ccc}
\toprule
    Question & \thead{Visco--elastic\\ control} & \thead{Behavioural\\
    maps control}  \\
    \midrule
    \makecell{The experience\\ with the walker\\ was pleasant?} & $3.38$
    - $0.92$ & $4.75$ - $0.46$  \\
    \midrule
    \makecell{You had the\\ impression you\\ had no control?} & $2.88$ -
    $0.83$ & $1.38$ - $0.74$  \\
    \midrule
    \makecell{The walker\\ hindered/prevented\\ your usual way of\\
    walking?} & $1.63$ - $0.74$ & $1.00$ - $0.00$  \\
    \midrule
\end{tabular}
\label{table:user-evaluation2}
\end{table}
The questions in Table~\ref{table:user-evaluation1} could be answered
with "yes" or "no". In the table, beside its relative question, there
is the percentage of answers that were "yes". The questions of
Table~\ref{table:user-evaluation2} could be answered with a value
between 1 and 5, were 1 means "not at all" and 5 means "extremely".
The aggregate result is presented as mean and standard deviation of
the answers.

\revv{From the results in Table~\ref{table:user-evaluation1} we can
  deduct that the Behavioural Maps' control is less intrusive, aiding
  the user's navigation without sacrificing her/his comfort. Through
  the questions reported in Table~\ref{table:user-evaluation2}, we
  could evaluate the cognitive aspects derived from the experience of
  using the walker.  We can observe a good level of accordance between
  the $12.5\%$ reported in both questions of
  Table~\ref{table:user-evaluation1} for the Behavioural map control
  and $1.38$ reported in the second question of
  Table~\ref{table:user-evaluation2}.  Likewise, the low performance
  reported for the visco-elastic control in the first question of
  Table~\ref{table:user-evaluation2} is an evident consequence of its
  perceived level of authority and intrusiveness reported in
  Table~\ref{table:user-evaluation1}. The evident conclusion is that
  the impression of being in control (at least in part) has an evident
  positive impact on the quality of the user's experience.}

Moreover $100\%$ of the participants preferred the control strategy
proposed in this paper over the classic visco--elastic control. Some
of the motivations were that our method gives more autonomy and
freedom to perform any path while the turns were performed more
softly, without forcing the participant to a particular trajectory.

\revv{In this section, we have shown a complete experimental
  evaluation both from the perspective of the quantitative performance
  and of the user experience.  In both cases, the results are very
  good and prove that this framework provide a navigation assistance,
  which guarantees a good level of agreement of the user trajectories
  with socially acceptable behaviours limiting at the same time the
  level of interference of the system with the user's choices.}


\section{Conclusions}
\label{sec:conclusions}

In this paper, we have considered a robot-assisted navigation
scenario. We adopted a shared authority controller, in which the
navigation decisions are shared between the human and the robot. The
key contribution of the paper is to show \emph{how} this decision can
be taken based on the degree of conformance of the human's behaviour
with the standard behaviour taken by humans in similar situation. We
substantiated this idea by a combination of learning and control
approaches, where the former are used to understand and classify the
human behaviours and the latter to change the visco-elastic parameters
of the guidance algorithm to adapt to the level of confidence that we
have on the human behaviour.

In the future, we will continue this research in many directions.
\revv{The most important one is removing the need for a prior
  knowledge of the map. A possible approach could be to study
  behavioural templates associated with specific features of the
  environment. During the execution, the system could classify the
  environment features and associate them on-the-fly with the expected
  behavioural templates.} \revv{We will also investigate other methods
  to increase the accuracy in distinguishing between turns and going
  straight. This could be done increasing the number of features or
  increasing the number of classes.}  \revv{The assumption that humans
  use clothoids as their preferred choice is certainly true in the
  majority of the cases and, especially, when the road in front of the
  human is sufficiently clear. In case these conditions are not met
  (e.g., a densely populated environment), the assumption could no
  longer hold. In the future we plan to analyse also these special
  cases. } \revv{In addition, it is possible that people deviate from
  this standard motion pattern because of cognitive or physical
  problems. In some of theses cases, using a clothoid as a reference
  trajectory could be seen as a rehabilitation policy.  These
  hypotheses needs further investigations.}  \revv{At last we want to
  evaluate what happens when the goal is changed dynamically during
  the online execution. This should not cause any abnormal behaviour
  in the control system: if the goal is altered while the user is in
  motion, the current motion may no longer align with the desired
  one. Consequently, the probability score for the desired motion
  class may decrease, leading to a change in the control
  hyperparameter. This adjustment increases the amount of control
  interventions, resulting in a more substantial elastic recall and a
  smoother correction of the current motion.}


\section*{Acknowledgements}
This work was supported by the European Union under NextGenerationEU
(FAIR - Future AI Research - PE00000013).


\bibliography{biblio}

\end{document}